\documentclass{article}

\usepackage{arxiv}
\usepackage[ruled,vlined]{algorithm2e}
\usepackage{tabularx}
\usepackage{url} 
\usepackage{multirow}
\usepackage{makecell}
\usepackage{microtype}
\usepackage{color, colortbl}
\usepackage[table]{xcolor}
\usepackage[colorinlistoftodos, textwidth=20mm]{todonotes}
\usepackage{mathtools}
\usepackage{eucal}
\usepackage{booktabs}
\usepackage{amsfonts}

\usepackage{amsmath}
\usepackage{caption,subcaption,graphicx}

\usepackage{tabularx}
\usepackage{multirow}
\usepackage{makecell}
\usepackage[table]{xcolor}
\usepackage{booktabs}
\usepackage{url}
\usepackage{tablefootnote}

\makeatletter
\renewcommand\@cite[2]{[{#1\if@tempswa\ (#2)\fi}]}
\makeatother

\usepackage{xcolor}
\usepackage{pgfplots}
\pgfplotsset{compat = 1.3} 
\usepackage{tikz}

\definecolor{bblue}{HTML}{4F81BD}
\definecolor{rred}{HTML}{C0504D}
\definecolor{ggreen}{HTML}{9BBB59}
\definecolor{ppurple}{HTML}{9F4C7C}

\def\x{{\mathbf x}}
\def\y{{\mathbf y}}

\title{Pseudo-Labeling for Massively Multilingual Speech Recognition}
\author{
    Loren Lugosch$^1$\thanks{Work done during an internship at FAIR. $^{\dagger}$Currently at Apple.} \quad
    Tatiana Likhomanenko$^{2\dagger}$ \quad Gabriel Synnaeve$^2$ \quad Ronan Collobert$^{2\dagger}$ \\
    $^1$McGill University / Mila, $^2$Facebook AI Research
}
\begin{document}
%

\maketitle
\begin{abstract}

Semi-supervised learning through pseudo-labeling has become a staple of state-of-the-art monolingual speech recognition systems.
In this work, we extend pseudo-labeling to massively multilingual speech recognition with 60 languages. We propose a simple pseudo-labeling recipe that works well even with low-resource languages: train a supervised multilingual model, fine-tune it with semi-supervised learning on a target language, generate pseudo-labels for that language, and train a final model using pseudo-labels for all languages, either from scratch or by fine-tuning. Experiments on the labeled Common Voice and unlabeled VoxPopuli datasets show that our recipe can yield a model with better performance for many languages that also transfers well to LibriSpeech.

\end{abstract}
\section{Introduction}
\label{sec:intro}

One of the long-term goals of automatic speech recognition (ASR) research is a single system that can transcribe speech in any language \cite{schultz1998multilingual, pratap2020massively}. Such a multilingual system would be simpler to maintain than a collection of monolingual models, enable users to comfortably speak any language without needing to tell the system which language to expect in advance, and share knowledge between all languages for improved performance.

A key ingredient of modern state-of-the-art monolingual ASR missing from current multilingual models
is \textit{pseudo-labeling} \cite{lee2013pseudo}, a technique for harnessing unlabeled datasets that has recently begun consistently yielding performance gains even for ASR tasks with large labeled datasets like LibriSpeech~\cite{synnaeve2019end,zhang2020pushing,xu2021self}. In pseudo-labeling, a model trained on a labeled dataset is used to generate labels for an unlabeled dataset, and those pseudo-labels (PLs) are then used to train a model. Many variants of pseudo-labeling exist: for instance, the same model used to generate PLs can also be trained on those PLs \cite{xu2020iterative,higuchi2021momentum,likhomanenko2021slimipl}, or PLs generated by a teacher model can be used to train a new student model \cite{synnaeve2019end,xu2021self, kahn2020self, park2020improved}.

\begin{figure}
    \centering
    \includegraphics{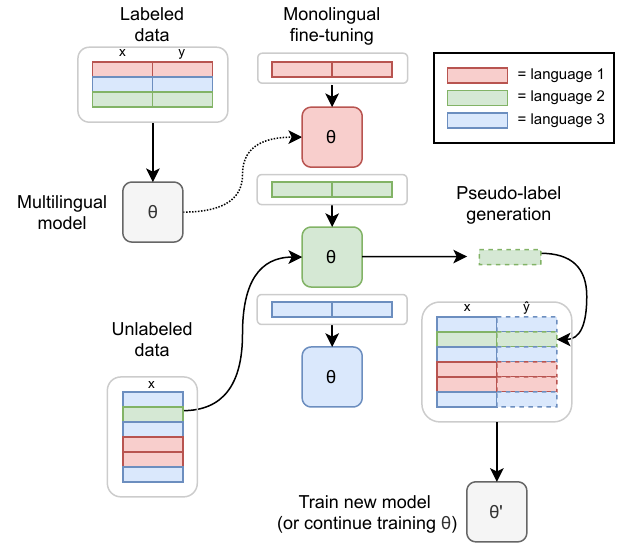}
    \caption{Illustration of our method: to produce better pseudo-labels for a given language, we first fine-tune the multilingual model on that language.}
    \label{training_diagram}
    \vspace{-0.25cm}
\end{figure}

In this work, we go beyond the monolingual setting and demonstrate the use of pseudo-labeling to improve a massively multilingual speech recognizer trained on all 60 languages of the Common Voice dataset~\cite{ardila2020common} simultaneously. First, we show that self-training on all unlabeled data in the multilingual VoxPopuli dataset~\cite{wang2021voxpopuli} at once tends to produce poor PLs for low-resource languages, and instead propose a simple recipe (Fig. \ref{training_diagram}) in which the model is first fine-tuned for a particular language before pseudo-labeling. Next, we compare a number of methods for training with the generated PLs, and find that training a larger model from scratch on all labeled and pseudo-labeled data, followed by fine-tuning on labeled data, works best. Finally, we show that the use of pseudo-labeled data improves out-of-domain generalization through experiments on LibriSpeech~\cite{panayotov2015librispeech}. Unlike much previous work on this topic,
our experiments use only open-source data, and we release our code and models for those who would like to experiment with them further.\footnote{For code, model checkpoints, and a Colab notebook showing how to perform inference, see: \url{https://github.com/flashlight/wav2letter/tree/main/recipes/mling_pl}}

\section{Model}
\begin{figure}
    \centering
    \includegraphics{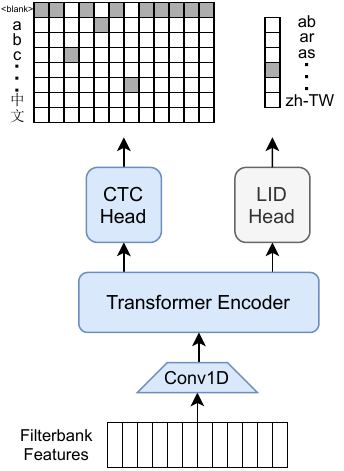}
    \caption{Illustration of the model used in our experiments, with optional language identification head (Sec. \ref{lid_section}) shown in grey.}
    \label{fig:model}
\end{figure}

The model used in our experiments
(Fig. \ref{fig:model}) 
is identical to the neural network used for LibriSpeech in \cite{likhomanenko2021slimipl}, except for the output layer(s). 
The input to the encoder is a sequence of 80-dimensional log mel filterbank frames, extracted using 25 ms Hamming windows every 10 ms from the 16 kHz audio signal.
The encoder has a single convolutional layer with a filter length of 7 and a stride of 3, followed by 36 transformer layers with 4 heads, feedforward dimension 3072, and self-attention dimension 768, using the relative position embeddings of \cite{shaw2018selfattention}.
The output of the encoder is fed to a CTC~\cite{graves2006connectionist} head and a language identification (LID) head. The CTC head is a linear layer with 8065 outputs: one for each character (most of which are Chinese characters), including punctuation, space, and the CTC \textless{}blank\textgreater{} symbol.
The CTC head is shared across all languages: it is a ``joint'' multilingual model, using the terminology of \cite{pratap2020massively}. The LID head is a linear layer with 60 outputs (one per language), followed by mean-pooling to aggregate the variable-length sequence of output vectors into a single vector of logits. The LID head outputs are only used during training: during inference, standard decoding algorithms can be applied to the CTC head outputs. The model is implemented and trained using Flashlight \cite{flashlight}.

While we do not perform explicit empirical comparisons with other multilingual models in the literature (as the focus of this work is on pseudo-labeling), it is worth noting that our model is significantly simpler than existing multilingual models, forgoing the use of language- or language-family-specific parameters, decoders, and tokenizers. We are not the first to use an encoder-only CTC architecture for multilingual ASR \cite{muller2017phonemic, tong2018cross, conneau2020unsupervised}, but we believe we are the first\footnote{We recently learned of an unpublished similar model based on XLSR-53  \cite{conneau2020unsupervised}: \url{https://t.co/yrr7pfuVVo} } to demonstrate this for \textit{massively} multilingual end-to-end ASR. 
Previous work on this topic \cite{adams2019massively, kannan2019large, cho2018multilingual, li2018bytes, hou2020large} has instead used more sophisticated sequence transduction models with autoregressive decoder networks
\cite{prabhavalkar2017comparison,chorowski2015attention, chan2016listen,graves2012sequence, he2019streaming}, 
citing the flaw of CTC's conditional independence assumption. 
In practice, CTC models implemented using modern neural network architectures are able to learn strong implicit language models~\cite{synnaeve2019end,likhomanenko2021slimipl} and achieve state-of-the-art results for the low-resource setting~\cite{baevski2020wav2vec,likhomanenko2021slimipl}. 
For those reasons, we focus on CTC models in this paper.

\vspace{-0.25cm}
\section{Data}\label{datasec}
The model is trained using the December 2020 release (6.1) of Common Voice (CV)~\cite{ardila2020common}, which has 3.6k hours of training data. CV is a continuously growing multilingual speech dataset recorded online by volunteer speakers. The 60 constituent languages vary greatly in the amount of available data: 7 languages have more than 100h of data,
and 10 languages have less than 1h of data.
We do not remove punctuation and capitalization from the CV transcripts, as this makes it easier to replicate our setup\footnote{While there have been attempts to standardize the formatting of transcripts for Common Voice for English~\cite{likhomanenko2020rethinking}, most reported results use an ad-hoc normalization scheme, and so cannot readily be compared.} and learning speed was not noticeably impacted. We downsample all audio to 16 kHz.

In addition to CV, we use VoxPopuli (VP) \cite{wang2021voxpopuli}, a very large scale (384k hours) unlabeled multilingual dataset of European languages. The dataset is split into 23 languages. 19 of the 23 VP languages are in CV (Czech, German, Greek, English, Spanish, Estonian, Finnish, French, Hungarian, Italian, Lithuanian, Latvian, Maltese, Dutch, Polish, Portuguese, Romanian, Slovenian, and Swedish): we use only those 19 languages for semi-supervised learning.

\section{Supervised training}\label{lid_section}

We train supervised models on CV for $\sim$500k updates. The hyperparameters and training procedure are identical to those used in \cite{likhomanenko2021slimipl}, except we use 2 SpecAugment~\cite{park2019specaugment} time masks instead of 10 (using 10 masks was found to cover too much of the shorter CV audio), and the learning rate is halved just once, at 250k updates.
We do not use the language balancing technique of \cite{kannan2019large, pratap2020massively} to sample languages evenly (which we found easily overfit to the low-resource languages), or curriculum learning 
as in~\cite{pratap2020massively}.
In addition to the base model (275M params), we also train larger models (1.06B params) by doubling the feedforward and self-attention dimensions of the transformer layers. The base models are trained on 16 GPUs with dynamic batching using 200s of audio per batch per GPU, and the large models are trained using 64 GPUs with 50s of audio per GPU, resulting in the same effective batch size.

\begin{figure}
    \centering
    \includegraphics{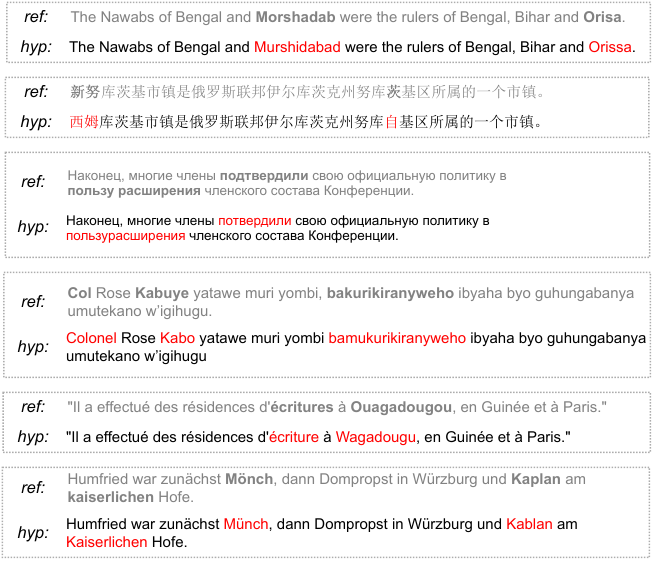}
    \caption{Example greedy decoding outputs from the base supervised model for 6 utterances from the validation sets of some of the higher-resource CV languages: English, Chinese (China), Russian, Kinyarwanda, French, and German.}
    \label{fig:high-resource-outputs}
\end{figure}

\begin{figure} 
    \centering
    \includegraphics{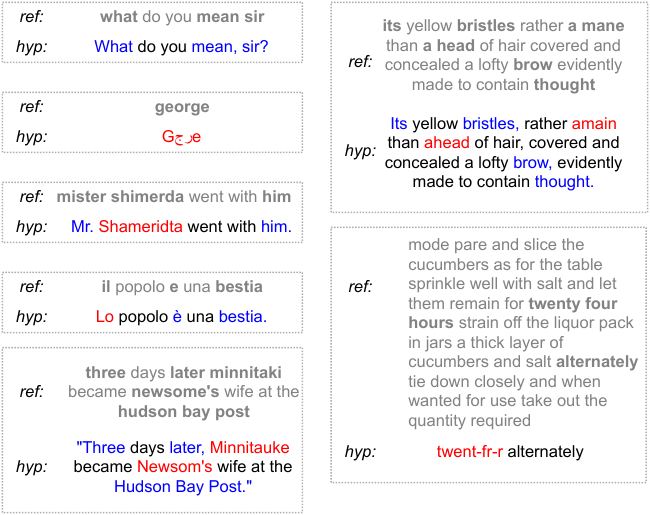}
    \caption{Examples of LibriSpeech dev-clean outputs with greedy decoding for base supervised model, trained only on CV, not on LibriSpeech. (Substitutions are colored: red = genuine error, blue = punctuation/truecasing counted as error.) Note that the model almost correctly transcribes the unusual Italian sentence in dev-clean, unlike a typical LibriSpeech model (cf. ~\cite[Table 12]{baevski2020wav2vec}).}
    \label{fig:librispeech_mistakes}
\end{figure}

Following \cite{toshniwal2018multilingual}, we add an LID loss, so that the loss $\ell$ used for training is
    $\ell = \ell_{\text{CTC}} + \gamma \cdot \ell_{\text{LID}},$
where $\ell_{\text{CTC}}$ represents the CTC loss,  $\ell_{\text{LID}}$ represents the LID loss (the cross-entropy between the LID head outputs and the one-hot language label for a given utterance), and $\gamma$ is a hyperparameter. We trained models on CV with $\gamma\in\{0, 0.1, 1, 10\}$: $\gamma = 1$ yielded the best results, with 2.6\% absolute improvement in average validation character error rate (CER) over the baseline with $\gamma=0$ (no LID), using greedy decoding. Some examples of greedy decoding outputs for the base supervised model are shown in Fig. \ref{fig:high-resource-outputs} and Fig. \ref{fig:librispeech_mistakes}.

\section{Semi-supervised training}\label{sec:ssl}
To train on the unlabeled data in VP,
we use slimIPL~\cite{likhomanenko2021slimipl}, 
an iterative approach
in which a model is trained for a number of updates on labeled data, followed by continuous training using labeled data and pseudo-labeled data stored in a dynamic cache which is periodically updated with pseudo-labels (PLs) re-generated by the current model state using greedy decoding without an external language model (LM). 
We use a cache size of 1000, replacement probability 0.1, and $\lambda = 10$ (ratio of unlabeled batches to labeled batches).

\subsection{Fine-tuning before pseudo-labeling}
\begin{figure}
    \centering
    \includegraphics{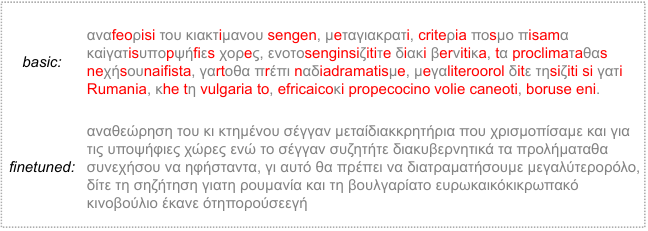}
    \caption{Pseudo-labels for an utterance from the Greek subset of VP with basic slimIPL (top) or with slimIPL after monolingual fine-tuning (bottom). Red letters are Latin characters.}
    \label{fig:greek_PLs}
    \vspace{-0.5cm}
\end{figure}

The simplest way to perform semi-supervised learning would be to pool the unlabeled data for all languages, as we do for the labeled data, and run slimIPL. We found that doing so led to poor PLs for low-resource languages, such as Greek, which has only 2.75h of training data (see top of Fig. \ref{fig:greek_PLs}~--- the transcript has a mix of Greek and Latin characters).

Instead, to produce PLs for a VP language, we first fine-tune the trained multilingual model by training only on CV data for that language for 10k updates, and then run slimIPL using the corresponding VP data (bottom of Fig. \ref{fig:greek_PLs}). The same effect could also be achieved
by generating PLs using a monolingual model,
but our proposed approach yields better results by taking advantage of multi-task learning (Table~\ref{greek_results_table}).

\begin{table}
\begin{center}
\caption{(Semi-)supervised learning results with slimIPL for the CV Greek data given different training sets.}
\label{greek_results_table}
\begin{tabular}{cccc}
\toprule
Labeled       & Unlabeled        & Valid CER & Test CER        \\
\midrule
CV All & ~-- & 53.2 & 47.8 \\
CV Greek  & ~--     &   30.6  &  33.6       \\ 
CV Greek  & VP Greek         & 23.9  &   25.1       \\
CV Greek   & VP English\tablefootnote{See Sec. \ref{wrong-lang} for a more detailed discussion of semi-supervised learning with language mismatch.}         & 24.3 &   28.4       \\
CV All $\to$ CV Greek     & ~--           & 9.9 &     9.6      \\
CV All $\to$ CV Greek     & VP Greek         & 8.7 &     8.5     \\
\bottomrule
\end{tabular}
\end{center}
\vspace{-0.5cm}
\end{table}

After training slimIPL models for all 19 languages in (CV languages~$\cap$~VP languages), we generate a final set of PLs\footnote{Pseudo-labels can be found at \url{https://dl.fbaipublicfiles.com/wav2letter/mling_pl/all_pseudo_labeled.lst} } for all unlabeled VP utterances using the appropriate slimIPL models. We filter out all utterances for which the PL length is 0 or \textgreater{}630 (maximum label length supported by the CTC loss implementation). The PLs for all languages can then be pooled and used either by continuing training the non-fine-tuned multilingual model checkpoint with all available CV and VP data, or by training a new model on that data from scratch. When training a model from scratch, we found it necessary for convergence to lower the learning rate from 0.03 to 0.01 and to delay SpecAugment until 50k updates; we also lower the learning rate when using VP data to fine-tune the base model already trained on CV.

Distilling the per-language fine-tuned models' knowledge back into a single final model is similar to the recently proposed multi-task self-training (MuST) \cite{ghiasi2021multi}. In MuST, a separate teacher model is trained for each task and used to pseudo-label every available training example, and a general student model with one head for each task is then trained on all the pseudo-labels. The difference here is that our final model only performs one ``task'', since we use a single shared CTC head over all languages, and the model itself must determine which language is being spoken.

\subsection{Avoiding collapse: cropping warmup period}\label{cropping}
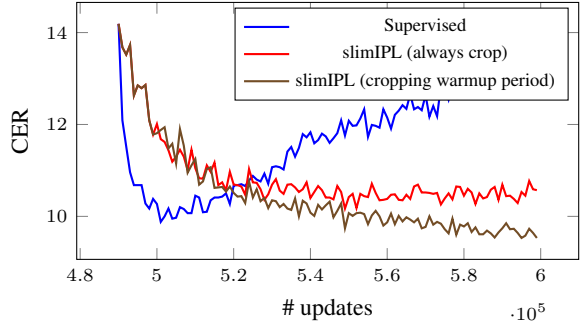
\begin{figure}
    \centering
    \begin{tikzpicture}
    \begin{axis}[
      xlabel={\small \# updates},
      ylabel={\small CER},
      width=0.5\columnwidth,
      height=5cm,
      ticklabel style={font=\scriptsize},
      legend style={font=\scriptsize},
      no markers,
      every axis plot/.append style={thick}
    ]
    \addplot table [y=cer, x=nupdate]{supervised_finetuning_greek.dat};
    \addplot table [y=cer, x=nupdate]{slimIPL_always_crop_greek.dat};
    \addplot table [y=cer, x=nupdate]{slimIPL_cropping_warmup_period.dat};
    \addlegendentry{Supervised}
    \addlegendentry{slimIPL (always crop)}
    \addlegendentry{slimIPL (cropping warmup period)}
    \end{axis}
    \end{tikzpicture}
    \caption{Validation CER for CV Greek (after training on CV All) with supervised fine-tuning or semi-supervised fine-tuning with VP Greek via slimIPL, using either a cropping warmup period or always cropping.}
    \label{fig:cropping_pl_performance}
    \vspace{-0.5cm}
\end{figure}

Another difficulty arose from the fact that the utterances of VP (average duration of 30s) are much longer than those of CV (average duration of 5.3s). The model trained only on CV generates mostly empty transcripts for VP, a commonly observed failure mode for out-of-domain audio or utterances longer than those observed during training~\cite{higuchi2021momentum, likhomanenko2021cape, chiu2021rnn}. 
Semi-supervised learning failed as a result, usually collapsing to generating all blanks even for the labeled data.
To acclimate the model to the longer VP utterances, we use a warmup period of 10k updates
during which we crop unlabeled audio into 10s segments before running the acoustic model, then stitch the resulting logit sequences back together and decode to obtain PLs. The model is then trained on the original uncropped utterance using those PLs.
Cropping the utterances results in poor pseudo-labels, so after a number of updates, we stop cropping the unlabeled utterances during pseudo-labeling. This warmup period approach works better than simply always cropping (Fig. \ref{fig:cropping_pl_performance}).

\begin{table}
\begin{center}
\caption{CER averaged over all CV languages.}
\label{all_results_table}
\begin{tabular}{lcc}
\toprule
Model      & Valid CER & Test CER \\
\midrule
Base model & 26.8 & 28.8 \\
+ all PLs (fine-tune) & 27.6 & 29.7 \\
+ all PLs (from scratch, base)  & 38.0 & 39.9 \\
$\hookrightarrow$ fine-tune on CV only  & 26.6 & 28.2 \\
+ all PLs (from scratch, large)  & 33.0 & 34.9 \\
$\hookrightarrow$ fine-tune on CV only  & 21.4 & 23.3 \\
\midrule
Monolingual baseline & 33.8 & 35.5 \\
Supervised fine-tuning & 10.6 & 11.4 \\
\bottomrule
\end{tabular}
\end{center}
\vspace{-0.5cm}
\end{table}

\vspace{-0.25cm}
\section{Performance on Common Voice}
\vspace{-0.25cm}

Table \ref{all_results_table} lists the performance of the multilingual model averaged over all CV languages in various settings.\footnote{Detailed per-language training logs and decoded outputs for all 60 languages can be found at \url{
https://dl.fbaipublicfiles.com/wav2letter/mling_pl/supplementary.zip}.} Table \ref{vp_results_table} reports the same information for CV languages that are in VP. All results for CV are reported using greedy decoding in terms of character error rate (CER), as suggested in~\cite{ardila2020common}. 

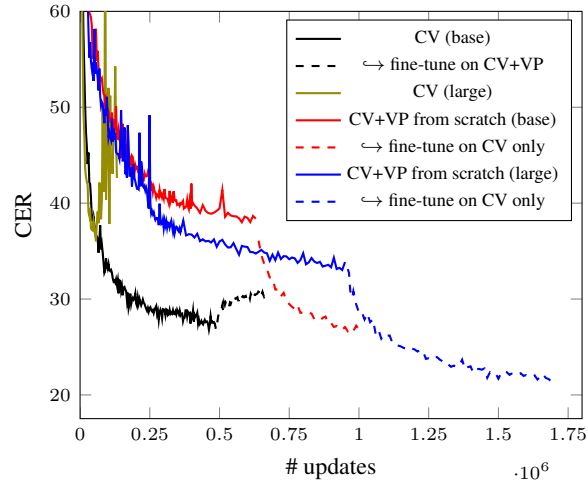
\begin{figure}[t]
    \centering
    \begin{tikzpicture}
    \begin{axis}[
      xlabel={\small \# updates},
      ylabel={\small CER},
      ymax=60,
      xmin=-100,
      xmax=1800000,
      xtick={0,250000,500000,750000,1000000,1250000,1500000,1750000},
      width=0.5\columnwidth,
      height=7cm,
      ticklabel style={font=\scriptsize},
      legend style={font=\scriptsize},
      no markers,
      every axis plot/.append style={thick}
    ]
    \addplot[color=black] table [y=cer, x=nupdate]{dev_cer_all_cv.dat};
    \addplot[color=black, dashed] table [y=cer, x=nupdate]{dev_cer_all_cv_then_cv_vp.dat};
    \addplot[color=olive] table [y=cer, x=nupdate]{dev_cer_all_cv_large.dat};
    \addplot[color=red] table [y=cer, x=nupdate]{dev_cer_all_cv_fromscratch.dat};
    \addplot[color=red,dashed] table [y=cer, x=nupdate]{dev_cer_all_cv_fromscratch_then_cv.dat};
    \addplot[color=blue] table [y=cer, x=nupdate]{dev_cer_all_cv_fromscratch_large.dat};
    \addplot[color=blue,dashed] table [y=cer, x=nupdate]{dev_cer_all_cv_fromscratch_large_then_cv.dat};
    \addlegendentry{CV (base)}
    \addlegendentry{$\hookrightarrow$ fine-tune on CV+VP}
    \addlegendentry{CV (large)}
    \addlegendentry{CV+VP from scratch (base)}
    \addlegendentry{$\hookrightarrow$ fine-tune on CV only}
    \addlegendentry{CV+VP from scratch (large)}
    \addlegendentry{$\hookrightarrow$ fine-tune on CV only}
    \end{axis}
    \end{tikzpicture}
    \caption{Validation CER curves for CV averaged over all languages for various training settings.}
    \label{fig:dev_curve_all}
    \vspace{-0.5cm}
\end{figure}

\begin{figure}[h]
\centering
\begin{subfigure}[b]{0.5\textwidth}
    \centering
    \begin{tikzpicture}[trim axis left, trim axis right]
    \begin{axis}[
      xlabel={\small \# updates},
      ylabel={\small CER},
      ymax=50,
      xmin=-100,
      xmax=1800000,
      xtick={0,250000,500000,750000,1000000,1250000,1500000,1750000},
      width=\columnwidth,
      height=7cm,
      ticklabel style={font=\scriptsize},
      legend style={font=\scriptsize},
      no markers,
      every axis plot/.append style={thick}
    ]
    \addplot[color=black] table [y=cer, x=nupdate]{dev_cer_vp_cv.dat};
    \addplot[color=black,dashed] table [y=cer, x=nupdate]{dev_cer_vp_cv_then_cv_vp.dat};
    \addplot[color=olive] table [y=cer, x=nupdate]{dev_cer_vp_cv_large.dat};
    \addplot[color=red] table [y=cer, x=nupdate]{dev_cer_vp_cv_fromscratch.dat};
    \addplot[color=red, dashed] table [y=cer, x=nupdate]{dev_cer_vp_cv_fromscratch_then_cv.dat};
    \addplot[color=blue] table [y=cer, x=nupdate]{dev_cer_vp_cv_fromscratch_large.dat};
    \addplot[color=blue,dashed] table [y=cer, x=nupdate]{dev_cer_vp_cv_fromscratch_large_then_cv.dat};
    \addlegendentry{CV (base)}
    \addlegendentry{$\hookrightarrow$ fine-tune on CV+VP}
    \addlegendentry{CV (large)}
    \addlegendentry{CV+VP from scratch (base)}
    \addlegendentry{$\hookrightarrow$ fine-tune on CV only}
    \addlegendentry{CV+VP from scratch (large)}
    \addlegendentry{$\hookrightarrow$ fine-tune on CV only}
    \end{axis}
    \end{tikzpicture}
    \caption{CV languages $\cap$ VP languages.}
    \label{fig:1}
\end{subfigure}\hfill
\begin{subfigure}[b]{0.5\textwidth}
    \centering
    \begin{tikzpicture}[trim axis left, trim axis right]
    \begin{axis}[
      xlabel={\small \# updates},
      ylabel={\small CER},
      ymax=80,
      xmin=-100,
      xmax=1800000,
      xtick={0,250000,500000,750000,1000000,1250000,1500000,1750000},
      width=\columnwidth,
      height=7cm,
      ticklabel style={font=\scriptsize},
      legend style={font=\scriptsize},
      no markers,
      every axis plot/.append style={thick}
    ]
    \addplot[color=black] table [y=cer, x=nupdate]{dev_cer_other_cv.dat};
    \addplot[color=black, dashed] table [y=cer, x=nupdate]{dev_cer_other_cv_then_cv_vp.dat};
    \addplot[color=olive] table [y=cer, x=nupdate]{dev_cer_other_cv_large.dat};
    \addplot[color=red] table [y=cer, x=nupdate]{dev_cer_other_cv_fromscratch.dat};
    \addplot[color=red,dashed] table [y=cer, x=nupdate]{dev_cer_other_cv_fromscratch_then_cv.dat};
    \addplot[color=blue] table [y=cer, x=nupdate]{dev_cer_other_cv_fromscratch_large.dat};
    \addplot[color=blue,dashed] table [y=cer, x=nupdate]{dev_cer_other_cv_fromscratch_large_then_cv.dat};
    \addlegendentry{CV (base)}
    \addlegendentry{$\hookrightarrow$ fine-tune on CV+VP}
    \addlegendentry{CV (large)}
    \addlegendentry{CV+VP from scratch (base)}
    \addlegendentry{$\hookrightarrow$ fine-tune on CV only}
    \addlegendentry{CV+VP from scratch (large)}
    \addlegendentry{$\hookrightarrow$ fine-tune on CV only}
    \end{axis}
    \end{tikzpicture}
    \caption{CV languages \textbackslash VP languages.}
    \label{fig:2}
\end{subfigure}
\caption{Validation CER curves for CV when averaging over the subset of languages in VP (left) and the subset of languages \textit{not} in VP (right).}
\label{fig:dev_curves_subsets}
\vspace{-0.5cm}
\end{figure}
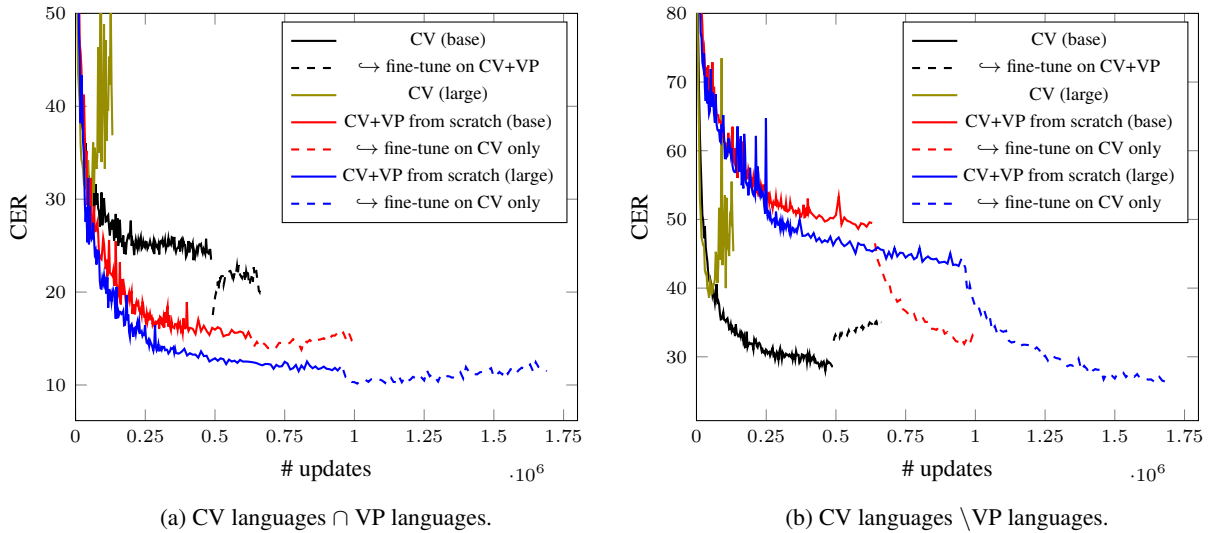

In addition to the base model (trained only on CV), we report performance when the VP audio with the final set of PLs is added back into the training set, either by fine-tuning the model already trained on CV (``+ all PLs (fine-tune)'') or by training a model from scratch on CV+VP (``+ all PLs (from scratch)''). We only report results for the large model when training it from scratch on CV+VP, as the large model overfit to CV after a few epochs (see Fig.~\ref{fig:dev_curve_all}, ``CV (large)''). Test CER is measured by selecting the checkpoint with the best average validation CER across all languages. While performance is degraded on average (Fig.~\ref{fig:dev_curve_all}), it is greatly improved for the VP languages (Fig.~\ref{fig:1}), with the best results achieved training a larger model from scratch.

The degradation for CV languages on average can be explained by the fact that VP is much larger than CV, leading to an imbalance in favor of the 19 languages in (CV languages~$\cap$~VP languages). If we then fine-tune the models trained on CV+VP on \textit{only} CV (``$\hookrightarrow$ fine-tune on CV only''), they not only still have improved performance over the base model when averaging over the VP languages, but also close the gap when averaging over all CV languages.

We also train a monolingual model for each language separately using the same hyperparameters as the multilingual model, and report the performance of those models along with the performance of the multilingual model when fine-tuned using only labeled data for that language (``supervised fine-tuning'') or, when unlabeled data is available (Table \ref{vp_results_table}), using both labeled and unlabeled data for that language (``slimIPL fine-tuning''). For monolingual models, or multilingual models with monolingual fine-tuning, the test CER is measured using the checkpoint with the best validation CER. There is still a large gap between the base model and fine-tuned models
(see e.g. Greek
in Table \ref{greek_results_table}), but the gap is reduced for the VP languages when training on the pseudo-labeled data.

\begin{table} 
\begin{center}
\caption{CER averaged over languages in (CV languages $\cap$ VP languages).}
\label{vp_results_table}
\begin{tabular}{lcc}
\toprule
Model      & Valid CER & Test CER        \\
\midrule
Base model & 24.4  & 24.8   \\
+ all PLs (fine-tune) & 17.5 & 17.9 \\
+ all PLs (from scratch, base)  & 15.0 & 15.6 \\
$\hookrightarrow$ fine-tune on CV only  & 13.8 & 14.0 \\
+ all PLs (from scratch, large)  & 11.5 & 12.0  \\
$\hookrightarrow$ fine-tune on CV only  & 10.1 & 10.6 \\
\midrule
Monolingual baseline  &  25.1   & 26.8 \\
Supervised fine-tuning & 7.7 & 8.3  \\
slimIPL fine-tuning   & 6.9 & 7.5   \\
\bottomrule
\end{tabular}
\end{center}
\vspace{-0.5cm}
\end{table}

\clearpage

\begin{figure}
\centering
\begin{subfigure}[b]{0.33\textwidth}
    \centering
    \begin{tikzpicture}[trim axis left, trim axis right]
    \begin{axis}[
      xlabel={\small \# updates},
      ylabel={\small CER},
      ymax=100,
      xmin=-100,
      xmax=1800000,
      xtick={0,250000,500000,750000,1000000,1250000,1500000,1750000},
      width=\columnwidth,
      height=6cm,
      ticklabel style={font=\scriptsize},
      legend style={font=\scriptsize},
      no markers,
      every axis plot/.append style={thick}
    ]
    \addplot[color=black] table [y=cer, x=nupdate]{dev_cer_greek.dat};
    \addplot[color=black,dashed] table [y=cer, x=nupdate]{dev_cer_greek_then_cv_vp.dat};
    \addplot[color=red] table [y=cer, x=nupdate]{dev_cer_greek_fromscratch.dat};
    \addplot[color=red, dashed] table [y=cer, x=nupdate]{dev_cer_greek_fromscratch_then_cv.dat};
    \addplot[color=blue] table [y=cer, x=nupdate]{dev_cer_greek_fromscratch_large.dat};
    \addplot[color=blue, dashed] table [y=cer, x=nupdate]{dev_cer_greek_fromscratch_large_then_cv.dat};
    \addlegendentry{CV}
    \addlegendentry{$\hookrightarrow$ CV+VP}
    \addlegendentry{CV+VP}
    \addlegendentry{$\hookrightarrow$ CV}
    \addlegendentry{CV+VP (large)}
    \addlegendentry{$\hookrightarrow$ CV}
    \end{axis}
    \end{tikzpicture}
    \caption{Greek --- 2.75h}
    \label{greek}
\end{subfigure}\hfill
\begin{subfigure}[b]{0.33\textwidth}
    \centering
    \begin{tikzpicture}[trim axis left, trim axis right]
    \begin{axis}[
      xlabel={\small \# updates},
      ylabel={\small CER},
      ymax=60,
      xmin=-100,
      xmax=1800000,
      xtick={0,250000,500000,750000,1000000,1250000,1500000,1750000},
      width=\columnwidth,
      height=6cm,
      ticklabel style={font=\scriptsize},
      legend style={font=\scriptsize},
      no markers,
      every axis plot/.append style={thick}
    ]
    \addplot[color=black] table [y=cer, x=nupdate]{dev_cer_finnish.dat};
    \addplot[color=black,dashed] table [y=cer, x=nupdate]{dev_cer_finnish_then_cv_vp.dat};
    \addplot[color=red] table [y=cer, x=nupdate]{dev_cer_finnish_fromscratch.dat};
    \addplot[color=red, dashed] table [y=cer, x=nupdate]{dev_cer_finnish_fromscratch_then_cv.dat};
    \addplot[color=blue] table [y=cer, x=nupdate]{dev_cer_finnish_fromscratch_large.dat};
    \addplot[color=blue, dashed] table [y=cer, x=nupdate]{dev_cer_finnish_fromscratch_large_then_cv.dat};
    \addlegendentry{CV}
    \addlegendentry{$\hookrightarrow$ CV+VP}
    \addlegendentry{CV+VP}
    \addlegendentry{$\hookrightarrow$ CV}
    \addlegendentry{CV+VP (large)}
    \addlegendentry{$\hookrightarrow$ CV}
    \end{axis}
    \end{tikzpicture}
    \caption{Finnish --- 0.55h}
    \label{finnish}
\end{subfigure}\hfill
\begin{subfigure}[b]{0.33\textwidth}
    \centering
    \begin{tikzpicture}[trim axis left, trim axis right]
    \begin{axis}[
      xlabel={\small \# updates},
      ylabel={\small CER},
      ymax=100,
      xmin=-100,
      xmax=1800000,
      xtick={0,250000,500000,750000,1000000,1250000,1500000,1750000},
      width=\columnwidth,
      height=6cm,
      ticklabel style={font=\scriptsize},
      legend style={font=\scriptsize},
      no markers,
      every axis plot/.append style={thick}
    ]
    \addplot[color=black] table [y=cer, x=nupdate]{dev_cer_lith.dat};
    \addplot[color=black,dashed] table [y=cer, x=nupdate]{dev_cer_lith_then_cv_vp.dat};
    \addplot[color=red] table [y=cer, x=nupdate]{dev_cer_lith_fromscratch.dat};
    \addplot[color=red, dashed] table [y=cer, x=nupdate]{dev_cer_lith_fromscratch_then_cv.dat};
    \addplot[color=blue] table [y=cer, x=nupdate]{dev_cer_lith_fromscratch_large.dat};
    \addplot[color=blue, dashed] table [y=cer, x=nupdate]{dev_cer_lith_fromscratch_large_then_cv.dat};
    \addlegendentry{CV}
    \addlegendentry{$\hookrightarrow$ CV+VP}
    \addlegendentry{CV+VP}
    \addlegendentry{$\hookrightarrow$ CV}
    \addlegendentry{CV+VP (large)}
    \addlegendentry{$\hookrightarrow$ CV}
    \end{axis}
    \end{tikzpicture}
    \caption{Lithuanian --- 1.18h}
    \label{lithuanian}
\end{subfigure}
\caption{Validation CER curves for the base and large multilingual models' performance on three low-resource CV languages with a corresponding subset in VP.}
\label{fig:dev_curves_low_resource_langs}
\end{figure}
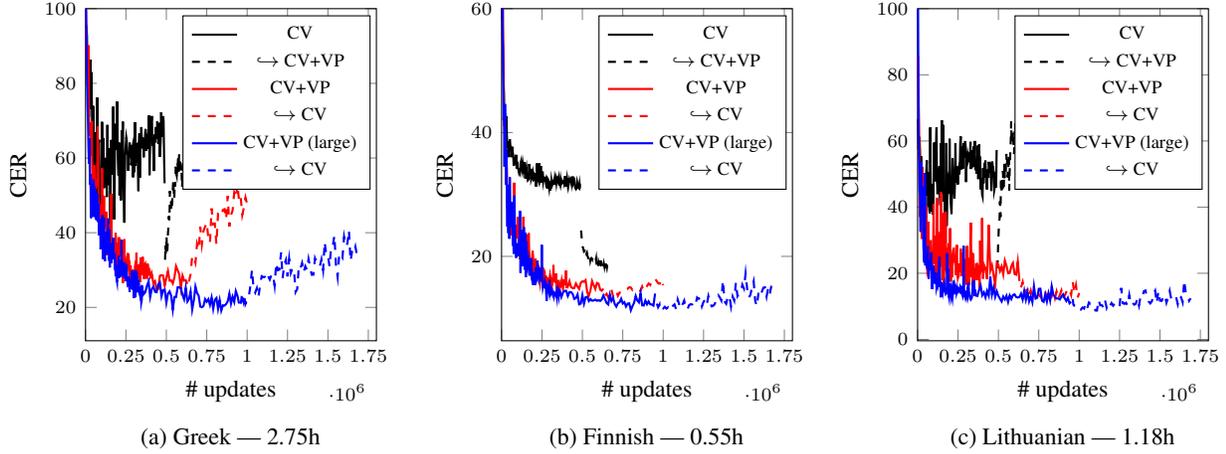

\begin{figure}
\centering
\begin{subfigure}[b]{0.33\textwidth}
    \centering
    \begin{tikzpicture}[trim axis left, trim axis right]
    \begin{axis}[
      xlabel={\small \# updates},
      ylabel={\small CER},
      ymax=45,
      xmin=-100,
      xmax=1800000,
      xtick={0,250000,500000,750000,1000000,1250000,1500000,1750000},
      width=\columnwidth,
      height=6cm,
      ticklabel style={font=\scriptsize},
      legend style={font=\scriptsize},
      no markers,
      every axis plot/.append style={thick}
    ]
    \addplot[color=black] table [y=cer, x=nupdate]{dev_cer_english.dat};
    \addplot[color=black,dashed] table [y=cer, x=nupdate]{dev_cer_english_then_cv_vp.dat};
    \addplot[color=red] table [y=cer, x=nupdate]{dev_cer_english_fromscratch.dat};
    \addplot[color=red, dashed] table [y=cer, x=nupdate]{dev_cer_english_fromscratch_then_cv.dat};
    \addplot[color=blue] table [y=cer, x=nupdate]{dev_cer_english_fromscratch_large.dat};
    \addplot[color=blue, dashed] table [y=cer, x=nupdate]{dev_cer_english_fromscratch_large_then_cv.dat};
    \addlegendentry{CV}
    \addlegendentry{$\hookrightarrow$ CV+VP}
    \addlegendentry{CV+VP}
    \addlegendentry{$\hookrightarrow$ CV}
    \addlegendentry{CV+VP (large)}
    \addlegendentry{$\hookrightarrow$ CV}
    \end{axis}
    \end{tikzpicture}
    \caption{English --- 897h}
    \label{english}
\end{subfigure}\hfill
\begin{subfigure}[b]{0.33\textwidth}
    \centering
    \begin{tikzpicture}[trim axis left, trim axis right]
    \begin{axis}[
      xlabel={\small \# updates},
      ylabel={\small CER},
      ymax=30,
      xmin=-100,
      xmax=1800000,
      xtick={0,250000,500000,750000,1000000,1250000,1500000,1750000},
      width=\columnwidth,
      height=6cm,
      ticklabel style={font=\scriptsize},
      legend style={font=\scriptsize},
      no markers,
      every axis plot/.append style={thick}
    ]
    \addplot[color=black] table [y=cer, x=nupdate]{dev_cer_catalan.dat};
    \addplot[color=black,dashed] table [y=cer, x=nupdate]{dev_cer_catalan_then_cv_vp.dat};
    \addplot[color=red] table [y=cer, x=nupdate]{dev_cer_catalan_fromscratch.dat};
    \addplot[color=red, dashed] table [y=cer, x=nupdate]{dev_cer_catalan_fromscratch_then_cv.dat};
    \addplot[color=blue] table [y=cer, x=nupdate]{dev_cer_catalan_fromscratch_large.dat};
    \addplot[color=blue, dashed] table [y=cer, x=nupdate]{dev_cer_catalan_fromscratch_large_then_cv.dat};
    \addlegendentry{CV}
    \addlegendentry{$\hookrightarrow$ CV+VP}
    \addlegendentry{CV+VP}
    \addlegendentry{$\hookrightarrow$ CV}
    \addlegendentry{CV+VP (large)}
    \addlegendentry{$\hookrightarrow$ CV}
    \end{axis}
    \end{tikzpicture}
    \caption{Catalan --- 444h}
    \label{catalan}
\end{subfigure}\hfill
\begin{subfigure}[b]{0.33\textwidth}
    \centering
    \begin{tikzpicture}[trim axis left, trim axis right]
    \begin{axis}[
      xlabel={\small \# updates},
      ylabel={\small CER},
      ymax=100,
      xmin=-100,
      xmax=1800000,
      xtick={0,250000,500000,750000,1000000,1250000,1500000,1750000},
      width=\columnwidth,
      height=6cm,
      ticklabel style={font=\scriptsize},
      legend style={font=\scriptsize},
      no markers,
      every axis plot/.append style={thick}
    ]
    \addplot[color=black] table [y=cer, x=nupdate]{dev_cer_kab.dat};
    \addplot[color=black,dashed] table [y=cer, x=nupdate]{dev_cer_kab_then_cv_vp.dat};
    \addplot[color=red] table [y=cer, x=nupdate]{dev_cer_kab_fromscratch.dat};
    \addplot[color=red, dashed] table [y=cer, x=nupdate]{dev_cer_kab_fromscratch_then_cv.dat};
    \addplot[color=blue] table [y=cer, x=nupdate]{dev_cer_kab_fromscratch_large.dat};
    \addplot[color=blue, dashed] table [y=cer, x=nupdate]{dev_cer_kab_fromscratch_large_then_cv.dat};
    \addlegendentry{CV}
    \addlegendentry{$\hookrightarrow$ CV+VP}
    \addlegendentry{CV+VP}
    \addlegendentry{$\hookrightarrow$ CV}
    \addlegendentry{CV+VP (large)}
    \addlegendentry{$\hookrightarrow$ CV}
    \end{axis}
    \end{tikzpicture}
    \caption{Kabyle --- 109h}
    \label{kab}
\end{subfigure}
\caption{Validation CER curves for the base and large  multilingual models' performance on three high-resource CV languages: English ($\in$ VP), Catalan ($\not\in$ VP), Kabyle ($\not\in$ VP).}
\label{fig:dev_curves_high_resource_langs}
\end{figure}
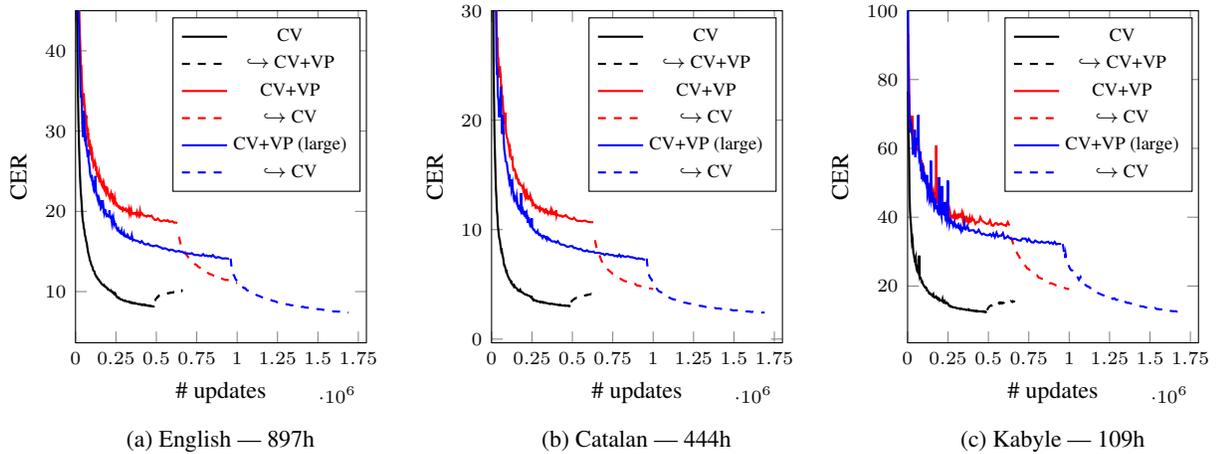

Fig. \ref{fig:dev_curves_low_resource_langs} shows the performance of the multilingual model being improved by pseudo-labeled data for three low-resource CV languages. In Fig. \ref{fig:dev_curves_high_resource_langs}, for three high-resource languages, performance is worse when fine-tuning on CV+VP and much worse when training a new model from scratch on CV+VP, but the performance gap is closed by fine-tuning the larger model on CV only.

There is a straightforward explanation for why the model trained from scratch on CV+VP initially performs so much worse on Catalan and Kabyle, before the model is fine-tuned only on CV: those languages are not in VP, so the amount of training data observed by the model for those languages is dwarfed by the amount of training data observed for the VP languages. However, English is among the VP languages, so it is surprising that the performance of English is also worse for the model trained from scratch on CV+VP, and that performance becomes worse when the model trained on CV is fine-tuned on CV+VP. It is worth noting that VP data is somewhat noisy:
much of it is spoken by interpreters attempting to translate, in real-time, what is being said by another speaker in another language---sometimes stumbling over a word or repeating themselves. The domain mismatch between this type of speech, as opposed to the prompted speech in CV, may explain the performance gap. Even though performance is degraded for the English subset of CV, the use of VP data does improve the model's ability to process English in a new domain, as we show in the next section.

\section{Transferring to LibriSpeech}

To see how well the multilingual models perform on out-of-domain audio, we evaluate them on LibriSpeech in Table~\ref{librispeech_performance}. Word error rate (WER) is reported both using greedy decoding and using a beam search for
\begin{equation}
\underset{\y}{\operatorname{argmax}} ~ \log p_{\theta}(\y|\x) + \alpha \log p^{\text{LM}}(\y) + \beta |\y|,
\end{equation}
where $p_{\theta}(\y|\x)$ is the probability of transcript $\y$ given input audio $\x$ according to the acoustic model, $p^{\text{LM}}(\y)$ is the probability of $\y$ according to an external 4-gram word-level LM trained on the LibriSpeech LM corpus, $|\y|$ denotes the length of $\y$, and $\alpha,\beta$ are set using a small grid search on the dev sets.
We find that the multilingual model fine-tuned with all VP PLs performs much better on LibriSpeech across all settings. It can be seen from Table \ref{librispeech_split}, in which test-other is split by the duration of utterances, that the improvement is due mostly to the model's ability to process longer sequences acquired from training on the longer VP utterances (see Sec.~\ref{cropping}). 

We also demonstrate the base model's transfer capability by fine-tuning it either on the 100h or 960h subset of LibriSpeech (Table \ref{librispeech_performance}, ``CV $\to$ LS-\{100,960\}''). 
During fine-tuning, instead of 2 SpecAugment masks (Sec. \ref{lid_section}), we use 10 masks, as in \cite{likhomanenko2021slimipl}, which we found yielded better performance.
With fine-tuning on LibriSpeech, performance is greatly improved for the 100h setup over the 100h-only training, while with 960h performance is similar or slightly worse.
We have not yet made these comparisons for the CV+VP models, but our other results suggest that similar benefits may be observed.

\begin{table} 
\begin{center}
\caption{LibriSpeech WER for different training sets (275M parameter model).}
\label{librispeech_performance}
\begin{tabular}{cccccc} 
\toprule
\multirow{2}{*}{Data} & \multirow{2}{*}{LM} & \multicolumn{2}{c}{Dev WER} & \multicolumn{2}{c}{Test WER} \\
\cmidrule(lr){3-4} \cmidrule(lr){5-6}
 &  & clean        & other        & clean    & other        \\
  \midrule
 \multirow{2}{*}{CV} & - & 59.7 & 60.1 & 62.0 & 62.8 \\
 & 4-gram & 33.7 & 34.3 & 37.6 & 37.7 \\
  \midrule
 CV & - & 34.1 & 41.7 & 33.5 & 42.5 \\
 $\to$ CV+VP & 4-gram & 8.8 & 15.9 & 9.0 & 16.8 \\
 \midrule
 CV+VP & - & 39.7 & 47.9 & 39.0 & 49.4 \\
 $\to$ CV & 4-gram & 10.1 & 17.8 & 10.4 & 19.5 \\
 \midrule
 CV & - & 4.8 & 13.7 & 5.1 & 13.6 \\
 $\to$ LS-100 & 4-gram & 3.3 & 9.7 & 3.8 & 9.9 \\
 \midrule
CV & - & 3.0 & 7.5 & 3.1 & 7.4 \\
$\to$ LS-960 & 4-gram & 2.1 & 5.3 & 2.6 & 5.8 \\
\midrule
\midrule
\multirow{2}{*}{LS-100} & - & 6.2 & 16.8 & 6.2 & 16.8 \\
 & 4-gram & 4.1 & 12.4 & 4.5 & 12.7 \\
\midrule
\multirow{2}{*}{LS-960} & - & 2.7 & 6.8 & 2.8 & 6.9 \\
 & 4-gram & 2.0 & 5.1 & 2.6 & 5.7 \\
 \bottomrule
\end{tabular}
\end{center}
\vspace{-0.3cm}
\end{table}

\begin{table} 
\begin{center}
\caption{WERs for test-other split over audio duration.}
\label{librispeech_split}
\begin{tabular}{cccccc} 
\toprule
\multirow{2}{*}{Data} & \multirow{2}{*}{LM} & \multicolumn{4}{c}{Duration} \\
\cmidrule{3-6}  &  & \textless{}10s        &  10-15s      &  15-20s    & \textgreater{}20s   \\
\midrule
\multirow{2}{*}{CV} & - & 46.6 & 83.6 & 99.1 & 99.9 \\
& 4-gram & 15.6 & 54.7 & 93.3 & 98.7 \\
\midrule
CV & - & 43.5 & 38.6 & 41.3 & 47.7 \\
$\to$ CV+VP & 4-gram & 17.0 & 15.2 & 17.2 & 20.8 \\
\midrule
CV+VP & - & 48.2 & 47.3 & 52.8 & 63.5 \\
$\to$ CV & 4-gram & 17.9 & 18.8 & 23.0 & 33.3 \\
\bottomrule
\end{tabular}
\end{center}
\end{table}

\section{Pseudo-labeling the wrong language}\label{wrong-lang}

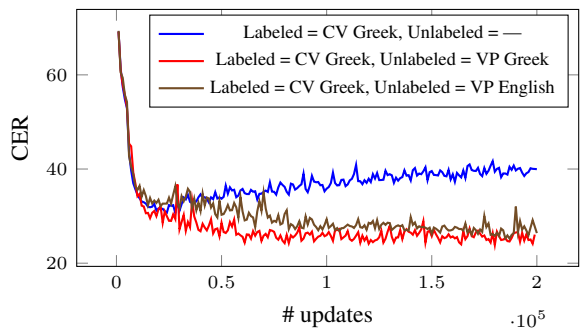
\begin{figure}
    \centering
    \begin{tikzpicture}
    \begin{axis}[
      xlabel={\small \# updates},
      ylabel={\small CER},
      width=0.5\columnwidth,
      height=5cm,
      ticklabel style={font=\scriptsize},
      legend style={font=\scriptsize},
      no markers,
      every axis plot/.append style={thick}
    ]
    \addplot table [y=cer, x=nupdate]{monolingual_cv_greek.dat};
    \addplot table [y=cer, x=nupdate]{monolingual_cv_greek_vp_greek.dat};
    \addplot table [y=cer, x=nupdate]{monolingual_cv_greek_vp_english.dat};
    \addlegendentry{Labeled = CV Greek, Unlabeled = ---}
    \addlegendentry{Labeled = CV Greek, Unlabeled = VP Greek}
    \addlegendentry{Labeled = CV Greek, Unlabeled = VP English}
    \end{axis}
    \end{tikzpicture}
    \caption{Validation CER for CV Greek for purely supervised monolingual training on CV Greek, using VP Greek as unlabeled data for slimIPL, or using VP English as unlabeled data.}
    \label{fig:language_mismatch_greek}
\end{figure}

To our surprise, we found that training a monolingual speech recognizer by pseudo-labeling the \textit{wrong language} could also improve test performance. Fig. \ref{fig:language_mismatch_greek} shows the validation CER of CV Greek when no unlabeled data, unlabeled data in the right language (VP Greek), and unlabeled data in the wrong language  (VP English) is used. 

This result may not currently be of much practical interest, since we can easily train a better monolingual Greek speech recognizer through other methods (Table \ref{greek_results_table}). Still, we believe it may be useful for understanding how and why semi-supervised learning works, and we hope to explore the phenomenon for more language pairs in the future.

\section{Conclusion}
We have demonstrated the use of pseudo-labeling to improve an end-to-end joint model for massively multilingual ASR with Common Voice. 
Fine-tuning a multilingual model with semi-supervised learning on each language of VoxPopuli separately,
 and then training on all VoxPopuli pseudo-labels combined, i) significantly improves the performance of the model for those 19 languages, ii) helps the model generalize to a new domain (LibriSpeech), and iii) enables training a larger model than was possible with Common Voice alone without overfitting.
Many interesting questions and problems remain, such as 
reducing the gap between the performance of the multilingual model on its own and after fine-tuning on a particular language, improving performance for languages without unlabeled data, integrating language models into the PL generation process, and running iterative pseudo-labeling instead of a single round with all languages. 
The method we have employed requires knowledge of which language is spoken in the unlabeled audio: overcoming this requirement, so that even more data in the wild can be used, would also be worth exploring.

\section*{Acknowledgments}
Thanks to Vineel Pratap for continuing training for the large models and for open-sourcing the pseudo-labels, code, and checkpoints.

\bibliographystyle{IEEEbib}
\bibliography{refs}

\end{document}